\documentclass[10pt,twocolumn,letterpaper]{article}

\usepackage[pagenumbers]{cvpr} 
\usepackage{pifont}
\usepackage{adjustbox}
\usepackage{makecell}
\usepackage{soul, color}
\usepackage{algorithm}

\newcommand{\xmark}{\ding{55}}

\definecolor{cvprblue}{rgb}{0.21,0.49,0.74}
\usepackage[pagebackref,breaklinks,colorlinks,allcolors=cvprblue]{hyperref}

\def\paperID{39789} 
\def\confName{CVPR}
\def\confYear{2026}

\title{E-TraMamba: A New Paradigm for Efficient Long-Term \\ 3D Feature Tracking with Event Cameras}

\author{Juwei Shen \\
  Department of Computing, FCMS\\
  The Hong Kong Polytechnic University\\
  Hong Kong, China \\
  {\tt\small sheldon.shen@connect.polyu.hk}
\and
Yujie Wu\\
  Department of Computing, FCMS\\
  The Hong Kong Polytechnic University\\
  Hong Kong, China \\
  {\tt\small yu-jie.wu@polyu.edu.hk} \\
\and
Changwen Chen\thanks{Corresponding Author}\\
  Department of Computing, FCMS\\
  The Hong Kong Polytechnic University\\
  Hong Kong, China \\
  {\tt\small changwen.chen@polyu.edu.hk} \\
}

\begin{document}
\maketitle
\begin{abstract}
Event-based 3D tracking enables low-latency and high-speed perception, while existing CNN- and Transformer-based trackers struggle to capture long-range spatiotemporal dependencies in sparse, noisy event streams, especially under real-time and efficiency constraints. To address these challenges, we present \textbf{E-TraMamba}, the first Mamba-based framework for 3D feature tracking on event data. This new framework adopts a linear state-space model for efficient long-range modeling and integrates a lightweight affine-transform predictor to maintain stable tracking under motion blur and occlusion. We also design an effective scheme to fuse multi-scale cues—local spatiotemporal patches, correlation maps, and positional embeddings—into a unified representation that enables stable and smooth 3D tracking. We construct a large-scale synthetic dataset, named \textbf{EvD-PointOdyssey}, which is generated with monocular rendering and provides synchronized event streams, depth maps, and accurate 3D trajectories for training and evaluating event-based 3D tracking models. Extensive experiments on event-based benchmarks demonstrate that E-TraMamba achieves state-of-the-art performance, delivering over \textbf{2$\times$ longer feature lifetimes} under strict accuracy thresholds (e.g., 0.1\,m), with higher tracked-feature ratios and lower RMSE than all baselines. These results make E-TraMamba a strong candidate for low-latency visual odometry, real-time SLAM, and interactive robotics.
\end{abstract}
    
\section{Introduction}
\label{sec:intro}

Feature tracking lies at the foundation of visual perception. It maintains the temporal consistency of salient and geometrically stable keypoints that support 3D reconstruction, SLAM, and motion estimation.
Unlike dense optical flow~\cite{ding2022eventopticalflow, Gehrig_2024eventopticalflow, eraftdenseopticalflow} and track-any-point tasks~\cite{xiao2024pointtrac, han2024Evpointtrac, liu2024EFpointtrac, karaev2024pointtrac, hamann2025Evpointtrac, wang2023framepoindttracking} that operate on dense motion fields, or high-level object tracking frameworks~\cite{yao2024mambamtrackuav, a74448884f0eventtrack, deng2025eventtrac, bd661f86ca32eventtrack, e7fbdafframeeventtrack, eb90145bd580eventtrack, tang2024frameeventtrac, 374af12d9bfframeeventtrac, 3b15935df4dframeeventtrac, a3601cc8b8f2194c6eventtrac, 722028431d78dfbebeventtrack, 901d1b5ad1e5a581eeventtrack, 6317b905d6ce1d5frameeventtrac, 5f470f7a4ad67f0e9frameeventtrac, 7bba0c5638bc1fd5frameeventtrack, 9e2fa156300220e64frameeventtrack, 9fabff24323e0d62f8bframeeventtrac, dad156b4f365dd32fa3d5eventtracksnntransformer, 11165142snneventmultitrac}, feature tracking focuses on sparse yet reliable trajectories that provide an interpretable geometric structure for long-term scene understanding. 
However, conventional frame-based pipelines~\cite{bay2006framefeatrack,323794framefeatrack,lowe2004framefeatrack,lucas1981framefeatrack,tomasi1991framefeatrack,ngo2025pointrac,kim2025pointtrac,xiao2024pointtrac,moing2024pointtrac,karaev2024pointtrac,wang2023framepoindttracking} inevitably degrade under challenging conditions such as motion blur and limited dynamic range, while their fixed frame rate introduces latency. These issues make them unreliable for many real-world tracking tasks, such as real-time robotics and high-speed perception.

\begin{figure}[t]
  \centering
   \includegraphics[width=1.0\linewidth]{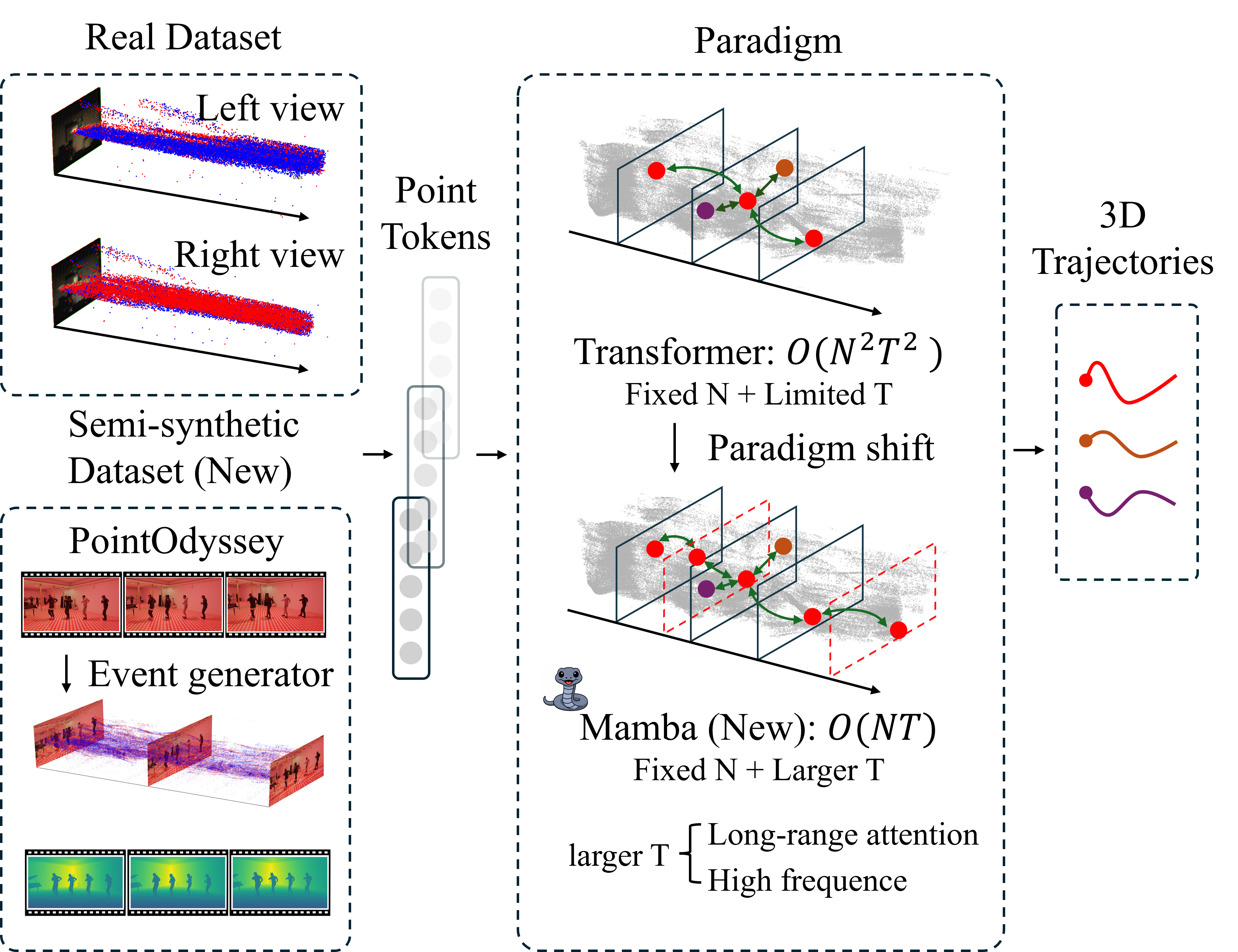}
    
    \caption{
    \textbf{Overview of the proposed E-TraMamba framework.}
    Our approach unifies real stereo event data (E-3DTrack) and a new semi-synthetic monocular+depth dataset (EvD-PointOdyssey) derived from PointOdyssey.
    Compared with Transformers requiring quadratic spatial–temporal attention $O(N^2T^2)$, 
    the Mamba-based formulation models feature evolution with linear complexity $O(NT)$, 
    enabling efficient long-range and high-frequency tracking of 3D trajectories.
    }
   \label{first-fig}
\end{figure}

A compelling alternative lies in event-based feature tracking, which leverages a fundamentally different sensing paradigm. Event cameras~\cite{Serreventcamera, 4444573eventcamera, 5648367eventcamera, 6889103eventcamera, 9897354eventcamera, 11043186eventcamera, dong2021eventcamera} asynchronously record per-pixel brightness changes with microsecond resolution and an extremely high dynamic range. This sparse data modality enables robust perception even in challenging scenarios characterized by fast motion and difficult lighting conditions~\cite{10377407eventenhance,10994378eventenhance,Liang_2025_ICCVeventillumination, 11165142snneventmultitrac, wang2025eventpingpong}. However, the asynchrony that provides these advantages breaks the dense and synchronous assumptions inherent in traditional vision systems~\cite{Mondal_2021_lackinf, 10.1109eventvisionsurvey_lackinfo}. This necessitates the development of new architectures capable of efficiently and continuously modeling long-range spatiotemporal dependencies.

Existing event-based feature tracking approaches can be divided into two main categories. \textbf{Event-assisted frameworks} combine the temporal precision of events with frame-based appearance cues to enhance robustness under fast motion and challenging illumination, but remain \emph{frame-dominant}, anchoring temporal reasoning to discrete frame intervals and resampling asynchronous data, which reintroduces latency and limits temporal fidelity. In contrast, \textbf{event-dominant frameworks} operate directly on asynchronous streams, preserving high temporal resolution for continuous, low-latency tracking. However, existing approaches~\cite{10077556Evpointtrac,3DEvpointtrac,7989517Evpointtrac,8491018Evpointtrac,hu2022Evpointtrac, han2024Evpointtrac, hamann2025Evpointtrac, mueggler2017Evpointtrac, alzugaray2020Evpointtrac} still face architectural bottlenecks: typical RNNs provide efficiency but limited memory, while Transformers capture global context at quadratic cost, restricting scalability for high-frequency data. As a result, event-based 3D feature tracking remains constrained by a trade-off between modeling capacity and computational efficiency.

To overcome these challenges, we design \textbf{E-TraMamba}, the first Mamba-based paradigm for event-based 3D feature tracking. E-TraMamba redefines feature tracking under a linear state-space formulation, enabling efficient long-range spatiotemporal modeling with linear complexity. It incorporates a lightweight affine-transform predictor for motion stability and fuses multi-scale cues—including spatiotemporal patches, correlation maps, and positional embeddings—into a unified representation for continuous 3D tracking. To address the scarcity of large-scale event data, we further construct \textbf{EvD-PointOdyssey}, a high-fidelity semi-synthetic dataset derived from PointOdyssey~\cite{zheng2023pointodysseydataset} with synchronized event streams, depth maps, and accurate 3D trajectories. Our contributions are threefold:

\begin{itemize}
    \item \textbf{New paradigm for event-driven tracking.} 
    We design the first Mamba-based framework for 3D feature tracking, establishing a new direction for efficient event-driven modeling.
    
    \item \textbf{Efficient long-term modeling for feature tracking.} 
    Built upon a structured state-space formulation, E-TraMamba explicitly models long-range spatiotemporal dependencies, achieving scalable and stable 3D tracking at high temporal frequencies.
    
    \item \textbf{The first high-fidelity synthetic dataset for event-based 3D feature tracking.}
    We construct \textbf{EvD-PointOdyssey}, a novel semi-synthetic benchmark featuring physically grounded event synthesis, realistic temporal dynamics, and unified evaluation for rigorous benchmarking.
\end{itemize}
\section{Related Work}
\subsection{Event-Assisted Feature Tracking}
To address the shortcomings of frame-based tracking under fast motion or adverse lighting, recent studies have explored frame–event fusion for keypoint detection and point tracking. Event cameras provide microsecond-level temporal cues that complement degraded or low frame-rate images, improving robustness and motion estimation~\cite{wang2024EFpointtrac,10629077EFpointtrac,shen2024EFpointtrac,liu2024EFpointtrac}. Despite their success, these fusion schemes remain constrained by the frame domain: the integration process still depends on discrete frame timestamps, which limits the temporal granularity and prevents full exploitation of the event stream’s asynchronous dynamics. These limitations highlight the need for \textbf{event-dominant frameworks} that operate natively in the event domain, enabling high-speed and long-term tracking.

\subsection{Event-Dominant Feature Tracking}
Contrary to fusion-based designs, \textbf{event-dominant frameworks} directly process asynchronous event streams, preserving their intrinsic temporal resolution and enabling continuous tracking under extreme motion or illumination. Early studies focused on handcrafted methods such as ICP alignment, patch warping, and event clustering~\cite{mueggler2017Evpointtrac,8491018Evpointtrac,7989517Evpointtrac}. More recent learning-based approaches—including DeepEvT~\cite{messikommer2023Evpointtrac}, Ev-TAP~\cite{han2024Evpointtrac}, and ETAP~\cite{hamann2025Evpointtrac}—demonstrate notable progress but still rely on synthetic video-to-event data and architectures such as RNNs or Transformers, which struggle to balance long-range modeling with efficiency. In contrast, our work \textbf{reframes the event-dominant paradigm} through a Mamba-based state-space formulation that achieves linear-complexity long-term reasoning, and introduces \textbf{EvD-PointOdyssey}, a high-fidelity semi-synthetic benchmark for unified 2D/3D evaluation.

\subsection{Mamba in Tracking Tasks}
Structured state-space models (SSMs) based on Mamba have recently been applied to tracking, but mainly at the object level. In multi-object tracking, Bi-Mamba~\cite{xiao2024mambamtrack} and related variants~\cite{hu2024mambamtrack,yao2024mambamtrackuav} model motion trajectories for robust MOT, while single-object trackers~\cite{li2024mambastracking,xie2024mambastracking,zhang2025mambastracking} emphasize long-context reasoning and modality fusion~\cite{liu2024mambaVLtracking,huang2024mambaVLtracking,wang2025mambaVLtracking,sun2025RGBEvtracking}. These models excel at object-level trajectory modeling with dense visual supervision, but have not addressed the fine-grained, geometry-aware domain of feature or point tracking, where sparse observations and asynchronous inputs pose distinct challenges. To our knowledge, this work is the first to extend the Mamba paradigm to event-driven 3D feature tracking.

\section{Methods}

\begin{figure*}[t]
\centering
\includegraphics[width=1.0\textwidth]{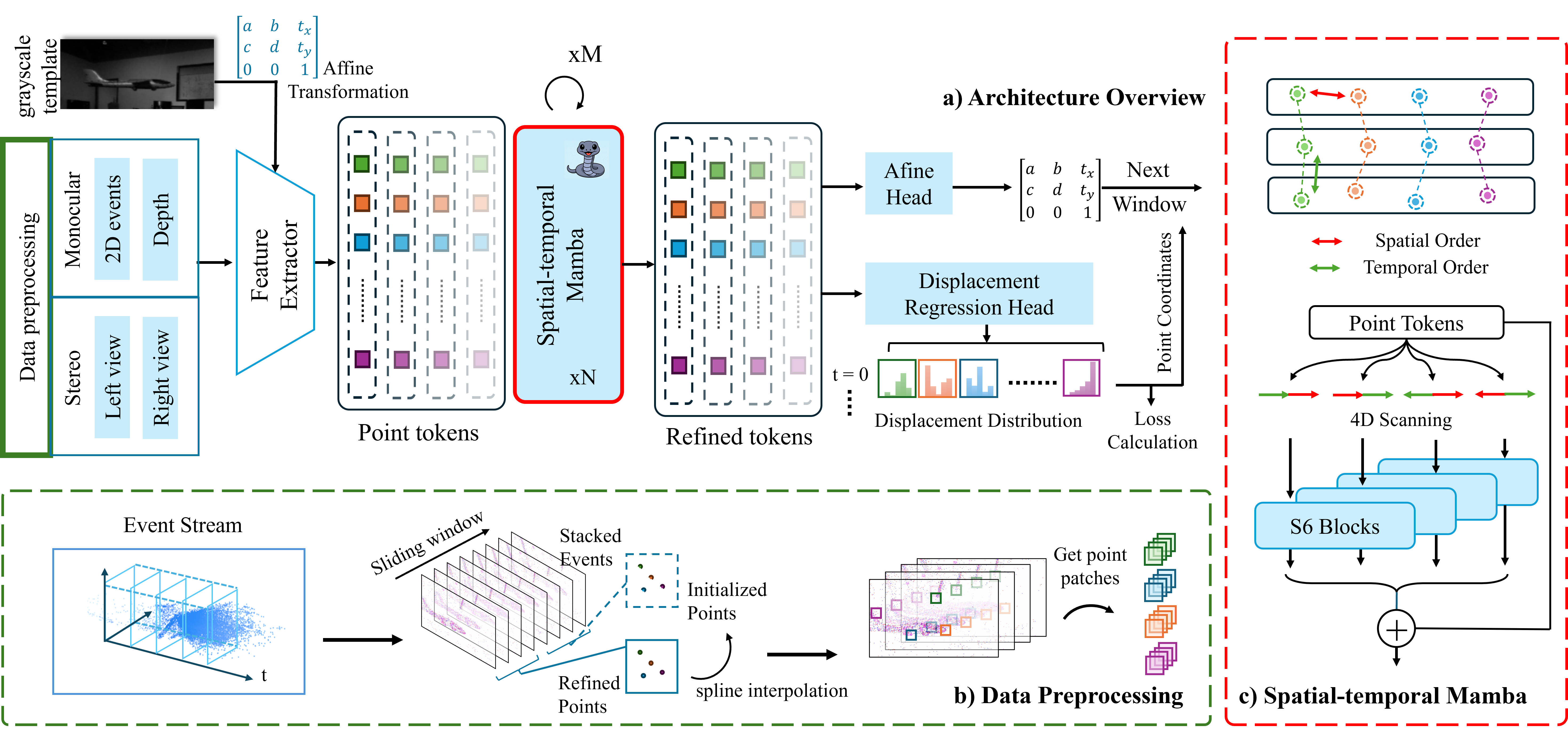}
\caption{
Overview of the proposed \textbf{E-TraMamba} framework.
\textbf{a) Architecture Overview:} the overall pipeline takes event streams (monocular or stereo) and optionally depth or grayscale templates as input. 
Extracted point tokens are refined through the Spatial-temporal Mamba module to model long-range dependencies. 
The affine and displacement regression heads predict per-point motion and affine transformations across time for window-to-window tracking. 
\textbf{b) Data Preprocessing:} raw event streams are organized into sliding windows and used to initialize and refine tracked points via spline interpolation. 
For each time window, local spatio-temporal patches (“Point Patchify”) are extracted around tracked locations to generate point tokens. 
\textbf{c) Spatial-temporal Mamba:} a 4D scanning mechanism processes tokens sequentially along both spatial and temporal orders, enabling efficient long-range modeling with lightweight S6 blocks.
}

\label{archfig}
\end{figure*}

\subsection{Preliminary}

\paragraph{Mamba.}
Mamba models sequential data using a state-space formulation, where the hidden state $\mathbf{z}[t]$ evolves over time under the influence of the input $\mathbf{x}[t]$. 
At each time step $t$, the continuous-time dynamics are defined as:
\begin{equation}
\mathbf{z}'[t] = \mathbf{A}\mathbf{z}[t] + \mathbf{B}\mathbf{x}[t], 
\qquad
\mathbf{y}[t] = \mathbf{C}\mathbf{z}[t].
\label{eq:continuous_ssm}
\end{equation}
Here, $\mathbf{A}$ denotes the system dynamics, while $\mathbf{B}$ and $\mathbf{C}$ are the input and output projection matrices. 
Using the zero-order hold (ZOH) method, the system is discretized as:
\begin{equation}
\widetilde{\mathbf{A}} = \exp(\Delta \mathbf{A}), 
\qquad 
\widetilde{\mathbf{B}} = \mathbf{A}^{-1}\!\big(\exp(\Delta \mathbf{A}) - \mathbf{I}\big)\mathbf{B},
\label{eq:zoh}
\end{equation}
leading to the discrete update rule:
\begin{equation}
\mathbf{z}[t] = \widetilde{\mathbf{A}}\mathbf{z}[t-1] + \widetilde{\mathbf{B}}\mathbf{x}[t], 
\qquad
\mathbf{y}[t] = \mathbf{C}\mathbf{z}[t].
\label{eq:discrete_update}
\end{equation}
Unlike classical state-space models, Mamba makes the parameters $\mathbf{B}$, $\mathbf{C}$, and $\Delta$ input-dependent:
\begin{equation}
(\mathbf{B},\, \mathbf{C},\, \Delta) = f_\theta(\mathbf{x}[t]),
\end{equation}
which allows dynamic transition behavior conditioned on the current token. 
A global convolution operator across time further refines $\mathbf{y}[t]$, 
enabling long-range temporal interaction with linear time and memory complexity. 
For detailed derivations, please refer to~\cite{gu2024mamba}.

\subsection{Model Architecture}

\paragraph{Track Token Representation.}
Following prior works~\cite{karaev2024pointtrac,xiao2024pointtrac,messikommer2023Evpointtrac,hamann2025Evpointtrac}, our framework represents each trajectory as a sequence of \emph{track tokens} that encode spatio-temporal and appearance information.
For a trajectory $i$, we define
\[
\mathbf{G}^i = [\mathbf{G}^i_1, \mathbf{G}^i_2, \dots, \mathbf{G}^i_T],
\]
where each token $\mathbf{G}^i_t$ aggregates multiple complementary features at the $t$-th time step.

Our model supports both stereo and monocular+depth configurations.
For \textbf{stereo 3D tracking}, we extract features from both left and right event views. 
For each tracked point, we compute: 
(1) local appearance features $\mathbf{F}^i_t$ from event patches via a CNN encoder, 
(2) correlation features $\mathbf{C}^i_t$ that measure temporal consistency between consecutive frames,
(3) reference template features $\mathbf{R}^i_t$ from an affine-aligned grayscale patch,
and (4) position encodings $\gamma_{\text{pos}}(\mathbf{x}^i_t)$ that encode spatial coordinates.
The left and right view features are then fused via an MLP:
\begin{align}
\mathbf{G}^{i,\text{L}}_t &= \mathrm{CNN}\big[\Omega^{\text{L}}_i(t)\big] + \mathbf{C}^i_t + \mathbf{R}^i_t + \gamma_{\text{pos}}(\mathbf{x}^i_t), \\
\mathbf{G}^{i,\text{R}}_t &= \mathrm{CNN}\big[\Omega^{\text{R}}_i(t)\big] + \mathbf{C}^i_t + \mathbf{R}^i_t + \gamma_{\text{pos}}(\mathbf{x}^i_t - \mathbf{d}_t), \\
\mathbf{G}^i_t &= \mathrm{MLP}_{\text{fusion}}\big(\mathbf{G}^{i,\text{L}}_t,\, \mathbf{G}^{i,\text{R}}_t\big),
\end{align}
where $\Omega^{\text{L/R}}_i(t)$ denotes the event patch and $\mathbf{d}_t$ is the disparity.

For \textbf{monocular+depth tracking}, depth information $\mathbf{D}^i_t$ is directly concatenated with the event data as an additional input channel before CNN encoding:
\begin{equation}
\mathbf{G}^i_t = \mathrm{CNN}\big[\Omega_i(t) \oplus \mathbf{D}^i_t\big] + \mathbf{C}^i_t + \mathbf{R}^i_t + \gamma_{\text{pos}}(\mathbf{x}^i_t),
\end{equation}
where $\oplus$ denotes channel concatenation.

This unified token representation compactly encodes geometric (position, depth/disparity, correlation) and appearance (local features, reference template) cues.
In the following sections, we primarily describe the stereo configuration for clarity, noting that the monocular+depth variant follows an analogous pipeline with depth maps replacing the right-view event stream.

\paragraph{Feature Extraction and Sliding Windows.}
Let  
\[
\mathcal{E}_{j}=\{(x_{k}, y_{k}, t_{k}, p_{k})\}_{k=1}^{N}
\]
denote the event stream collected between timestamps $[t_{j-1}, t_{j}]$, 
where $(x_{k}, y_{k})$ are pixel coordinates, $t_{k}$ the timestamp, and $p_{k}\in\{-1,+1\}$ the polarity. 
We process events using a sliding window strategy with window size $w$ and stride $w/2$:
\[
\mathcal{S}_{t} = \{(x, y, t', p) \in \mathcal{E} \mid t' \in [t-\tfrac{w}{2},\, t]\}.
\]
The half-window overlap ensures temporal continuity across adjacent windows.

For each tracked point $\mathbf{x}_{i}(t)$, we extract a square event patch  
\[
\Omega_{i}(t)=\{(u,v)\mid \lVert (u,v)-\mathbf{x}_{i}(t)\rVert_{\infty}\leq r\}.
\]
Point positions in the second half-window are initialized via cubic spline interpolation from the first half-window:
\[
\hat{\mathbf{x}}_{i}(t)=\mathrm{Spline}\big(\mathbf{x}_{i}(t-\tfrac{w}{2}),\dots\big).
\]

Each event patch is fed into a CNN encoder $\phi(\cdot)$ followed by global average pooling to produce local features $\mathbf{F}_{i}(t)$.
Correlation features $\mathbf{C}_{i}(t)$ capture temporal consistency by measuring patch similarity and 2D displacement between consecutive frames.
Reference template features $\mathbf{R}_{i}$ are extracted from a grayscale reference patch with learned affine alignment to handle scale and rotation changes.
All feature components are concatenated and fused through an MLP to produce the final token representation $\mathbf{G}^i_t \in \mathbb{R}^{d}$.
Detailed architecture specifications are provided in the supplementary material.

\paragraph{Mamba-based Refinement.}
Given the extracted track tokens $\mathbf{T}\!\in\!\mathbb{R}^{B \times T \times N \times d}$ 
(batch size $B$, time steps $T$, tracked points $N$, feature dimension $d$),
we apply iterative refinement through Mamba blocks to model long-range spatio-temporal dependencies.
Our architecture uses multiple refinement groups, each containing four directional Mamba modules that scan the token sequence along different orders.

The token tensor has two logical axes: the temporal axis ($\tau$) tracking each point across time, 
and the spatial axis ($\sigma$) relating different points at the same time step.
We construct four scanning directions: 
$\mathcal{S}=\{\tau^{+},\tau^{-},\sigma^{+},\sigma^{-}\}$, 
where $\tau^+$ and $\tau^-$ scan temporally (forward/backward), 
and $\sigma^+$ and $\sigma^-$ scan spatially (forward/backward).

For each direction $s\!\in\!\mathcal{S}$, we reshape the tokens into a 1D sequence via permutation $\pi_{s}$, 
process through a Mamba module, and inverse-permute back:
\[
\mathbf{H}_{s}
=\pi_{s}^{-1}\!\bigl(\mathrm{Mamba}_{s}\!\bigl(\pi_{s}(\mathbf{T})\bigr)\bigr).
\]
The outputs from all four directions are averaged and combined with the input via residual connection.
After iterative refinement, lightweight prediction heads decode the final tokens into 
2D positions (via softmax-weighted binning), disparity/depth values, and visibility confidence scores.

\subsection{Supervision and Loss Function}\label{supervision}
A unified multi-scale L1 supervision strategy is formulated to jointly constrain the 2D reprojection, disparity, and reconstructed 3D coordinates of each trajectory during training. Given predicted 2D points $\hat{\mathbf{x}}_{t}$, disparity $\hat{d}_t$, 
and reconstructed 3D coordinates $\hat{\mathbf{X}}_t$, 
with ground truths $\mathbf{x}_t$, $d_t$, and $\mathbf{X}_t$, 
the overall loss for a single refinement step is
\begin{equation}
\mathcal{L}_{\text{total}}
= \lambda_{\text{xy}} \| \hat{\mathbf{x}}_t - \mathbf{x}_t \|_1
+ \lambda_{\text{disp}} \| \hat{d}_t - d_t \|_1
+ \lambda_{\text{3D}} \| \hat{\mathbf{X}}_t - \mathbf{X}_t \|_1,
\end{equation}
where $\lambda_{\text{xy}}, \lambda_{\text{disp}}, \lambda_{\text{3D}}$ 
are balancing coefficients (set to 1.0). 
All losses are masked by visibility and averaged across all time steps and points. 
The total loss aggregates all refinement stages:
\begin{equation}
\mathcal{L}
= \sum_{k=1}^{K} \gamma^{K-k} \mathcal{L}_{\text{total}}^{(k)},
\end{equation}
where $\gamma$ is the decay factor (0.8). 
No adversarial or contrastive objectives are used—our supervision remains purely regression-based.

\subsection{Semi-Synthetic Dataset} \label{semi-syn}
We construct a high-fidelity semi-synthetic dataset, \textbf{EvD-PointOdyssey}, 
by extending the PointOdyssey framework~\cite{zheng2023pointodysseydataset}, 
which originally provides dense 3D point tracking sequences with precise geometry and long temporal coverage. 
While prior variants such as Ev-PointOdyssey~\cite{han2024Evpointtrac} simulate event data for dense motion supervision, 
they remain limited to dense tracking formulations. 
In contrast, our dataset redefines the paradigm toward \emph{sparse, event-based 3D feature tracking}, 
combining high-frame-rate RGB rendering with physics-based event simulation and per-frame depth supervision. 
We further apply strict temporal filtering and trajectory validation to ensure clean, geometrically consistent annotations, 
enabling large-scale training and benchmarking of event-based tracking models under realistic dynamics.
Given all candidate 3D points ${p_i^t}$, we retain only trajectories satisfying:
\begin{itemize}
\item Displacement constraint: maximum inter-frame displacement $\leq 12$ pixels;
\item Visibility constraint: each point remains visible throughout the sequence;
\item Depth-quality constraint: valid, finite depth values (no NaN or infinite entries);
\item Non-overlap constraint: selected temporal windows must not overlap;
\item Point-count constraint: up to 30 valid points per sequence.
\end{itemize}
This filtering eliminates noisy trajectories and yields stable, geometrically consistent tracks. 
The resulting dataset provides reliable event–depth–trajectory triplets, forming a solid foundation for benchmarking event-based 3D tracking models.

\section{Experiments}
\subsection{Dataset and Implementation Details}

\begin{figure}[t]
  \centering
   \includegraphics[width=0.8\linewidth]{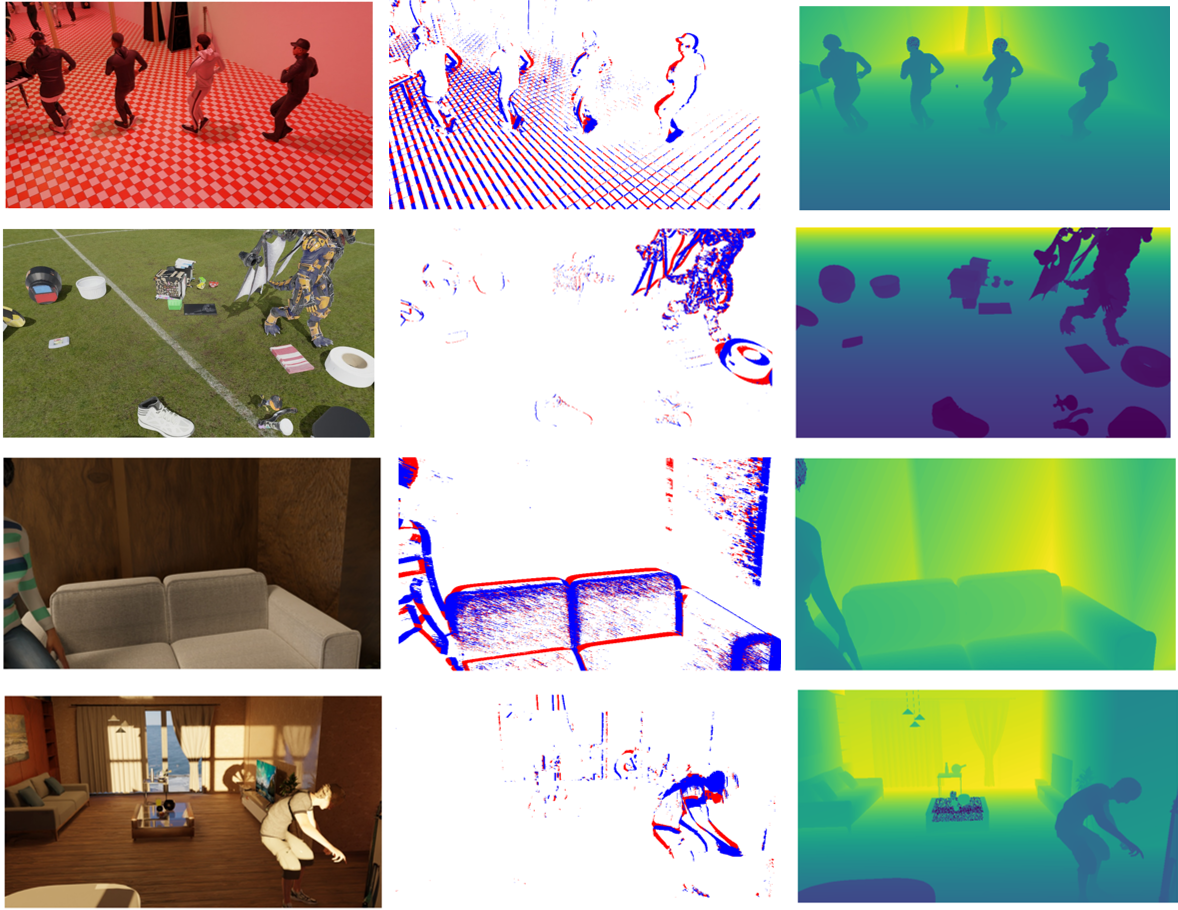}
    
   \caption{Examples from our generated semi-synthetic dataset showing corresponding RGB, event, and depth frames. The event image is obtained by simple accumulation of events within a \(\pm 0.03\) s window centered at each RGB timestamp.}
   \label{evd-po}
\end{figure}

\paragraph{Dataset.}
We evaluate our method on two datasets designed for event-based feature tracking, covering both 3D and 2D tasks. 
These include one large-scale real-world benchmark and one semi-synthetic dataset introduced in Section~\ref{semi-syn}.

1) \textbf{E-3DTrack}~\cite{3DEvpointtrac} is a stereo event-based dataset for 3D feature tracking, providing synchronized left–right event streams with ground-truth 3D feature trajectories across diverse motion patterns. It serves as a benchmark for evaluating long-range tracking stability and temporal consistency under real-world dynamic conditions.

2) \textbf{EvD-PointOdyssey}~(ours) is a semi-synthetic dataset derived from the PointOdyssey framework~\cite{zheng2023pointodysseydataset}, 
as detailed in Section~\ref{semi-syn}. 
Unlike E-3DTrack, it adopts a \emph{monocular+depth} configuration, pairing event streams with rendered depth maps to enable single-view 3D reasoning. 
For fair comparison with existing 2D trackers, evaluation is performed in the 2D projection domain. 
This design bridges 3D structure-aware tracking and conventional 2D event tracking. 
Representative examples from \textbf{EvD-PointOdyssey} are shown in Fig.~\ref{evd-po}.

\paragraph{Implementation Details.}

Our E-TraMamba model is built upon a Mamba-based architecture tailored for event-based feature tracking. We use a hidden dimension of 256 and a patch size of 31. The architecture consists of two Mamba refinement groups, each containing four directional Mamba blocks that process the sequence along both forward and backward directions across temporal and spatial dimensions. Each Mamba block implements a structured state-space model with a state dimension of 32, a convolutional kernel size of 4, and an expansion factor of 2. Our final model uses a window size of 12 to balance temporal modeling capacity and efficiency.

We apply positional encoding and dropout with a rate of 0.1. A two-stage refinement strategy is adopted, with a refinement decay factor of 0.8 to stabilize the optimization.

Training is conducted on 8 NVIDIA A6000 GPUs with a batch size of 16 for 120 epochs. We use the AdamW optimizer with an initial learning rate of $1 \times 10^{-4}$, a weight decay of $5 \times 10^{-4}$, and cosine annealing down to a minimum of $1 \times 10^{-6}$. Each training run takes approximately 2 days to complete.

\paragraph{Evaluation Metrics.}

To comprehensively evaluate our proposed E-TraMamba across both 3D and 2D event-based feature tracking tasks, 
we adopt metrics widely used in prior literature and tailored to the characteristics of each dataset.

For \textbf{E-3DTrack}~\cite{3DEvpointtrac}, which targets long-range and temporally consistent 3D tracking, 
we report the \emph{Tracked Feature Ratio (TFR)}, \emph{Feature Age (FA)}, and \emph{Root Mean Squared Error (RMSE)} under spatial thresholds of 0.1\,m, 0.15\,m, and 0.2\,m. 
TFR measures the proportion of trajectory duration during which the predicted 3D points remain within a predefined spatial error bound relative to ground truth, 
FA reflects the average lifetime of tracked features before loss, 
and RMSE quantifies the overall 3D deviation, providing a holistic measure of accuracy and stability.

\begin{table*}[t]
\centering
\begin{adjustbox}{width=\textwidth}
\begin{tabular}{l|ccc|ccc|c}
\toprule
\multicolumn{8}{c}{\textbf{E-3DTrack Results (3D evaluation)}} \\
\midrule
\textbf{Method} &
FA$_{(0.1\mathrm{m})}\,\uparrow$ & FA$_{(0.15\mathrm{m})}\,\uparrow$ & FA$_{(0.2\mathrm{m})}\,\uparrow$ &
TFR$_{(0.1\mathrm{m})}\,\uparrow$ & TFR$_{(0.15\mathrm{m})}\,\uparrow$ & TFR$_{(0.2\mathrm{m})}\,\uparrow$ &
RMSE (m)\,$\downarrow$ \\
\midrule
\makecell[l]{E-RAFT \cite{eraftdenseopticalflow} + TSES \cite{zhu2018eventstereodepth}} & 0.0409 & 0.0664 & 0.0920 & 0.1701 & 0.2667 & 0.3439 & 0.4726 \\
\makecell[l]{E-RAFT \cite{eraftdenseopticalflow} + SDE \cite{eventstereodepth}}         & 0.1385 & 0.2399 & 0.3204 & 0.3121 & 0.4726 & 0.5806 & 0.3368 \\
\makecell[l]{EKLT \cite{Gehrig_2018EKLTfeatracking} + TSES \cite{zhu2018eventstereodepth}} & 0.0232 & 0.0429 & 0.0628 & 0.1180 & 0.1961 & 0.2685 & 0.4806 \\
\makecell[l]{EKLT \cite{Gehrig_2018EKLTfeatracking} + SDE \cite{eventstereodepth}}     & 0.1026 & 0.1856 & 0.2584 & 0.2421 & 0.3738 & 0.4700 & 0.4034 \\
\makecell[l]{DeepEvT \cite{messikommer2023Evpointtrac} + TSES \cite{zhu2018eventstereodepth}} & 0.0713 & 0.1117 & 0.1452 & 0.3786 & 0.4991 & 0.5818 & 0.3549 \\
\makecell[l]{DeepEvT \cite{messikommer2023Evpointtrac} + SDE \cite{eventstereodepth}} & 0.2314 & 0.3462 & 0.4339 & 0.5782 & 0.7060 & 0.7765 & 0.1889 \\
E-3DTrack \cite{3DEvpointtrac} & 0.2601 & 0.4179 & 0.5428 & 0.6928 & 0.8164 & 0.8772 & 0.1181 \\
\textbf{E\textminus TraMamba (ours)} & \textbf{0.6146} & \textbf{0.7850} & \textbf{0.8657} & \textbf{0.7274} & \textbf{0.8440} & \textbf{0.9001} & \textbf{0.0914} \\
\midrule
\multicolumn{8}{c}{\textbf{EvD-PointOdyssey Results (2D evaluation)}} \\
\midrule
\textbf{Method} &
FA@2px~$\uparrow$ & FA@3px~$\uparrow$ & FA@5px~$\uparrow$ &
TFR@2px~$\uparrow$ & TFR@3px~$\uparrow$ & TFR@5px~$\uparrow$ &
RMSE-2D (px)~$\downarrow$ \\
\midrule
DeepEvT~\cite{messikommer2023Evpointtrac}
& 0.0867 & 0.1141 & 0.1693 & 0.0830 & 0.1087 & 0.1618 & 35.2671 \\
ETAP~\cite{hamann2025Evpointtrac} 
& \textbf{0.2179} & 0.2793 & 0.3782 & 0.1627 & 0.2204 & 0.3201 & 23.2615 \\
\textbf{E\textminus TraMamba (ours)} 
& 0.2177 & \textbf{0.3835} & \textbf{0.7094} & \textbf{0.2134} & \textbf{0.3782} & \textbf{0.7050} & \textbf{4.1625} \\
\bottomrule
\end{tabular}
\end{adjustbox}
\caption{
\textbf{Quantitative comparison on two event-based feature tracking benchmarks.}
The upper block reports \textbf{3D tracking results} on the \textbf{E-3DTrack} dataset~\cite{3DEvpointtrac}, 
evaluated under distance thresholds (0.1–0.2\,m) and 3D RMSE.
The lower block presents \textbf{2D tracking results} on the \textbf{EvD-PointOdyssey} dataset (ours),
evaluated under pixel thresholds (2–5\,px) and 2D RMSE.
E\textminus TraMamba consistently achieves the best accuracy and lowest error across both 3D and 2D evaluations.
}
\label{tab:combined_results_correct}
\end{table*}

For \textbf{EvD-PointOdyssey}, although our model is capable of predicting full 3D trajectories, 
most existing event-based trackers produce only 2D projections. 
To ensure a fair comparison, we therefore evaluate all methods in the 2D image plane following standard event-based feature tracking protocols. 
We report FA and TFR at pixel thresholds of 2\,px, 3\,px, and 5\,px, 
together with the \emph{RMSE-2D} that measures the average reprojection error across frames. 
TFR and FA respectively capture the success ratio and temporal stability of tracked features within each pixel threshold,
while RMSE-2D reflects the average spatial deviation of the projected points. 
The complete quantitative comparison is presented in Table~\ref{tab:combined_results_correct}. 
Together, these metrics jointly assess the precision, robustness, and cross-domain generalization of our approach in both 3D and 2D tracking settings.

\begin{figure}[t]
  \centering
   \includegraphics[width=\linewidth]{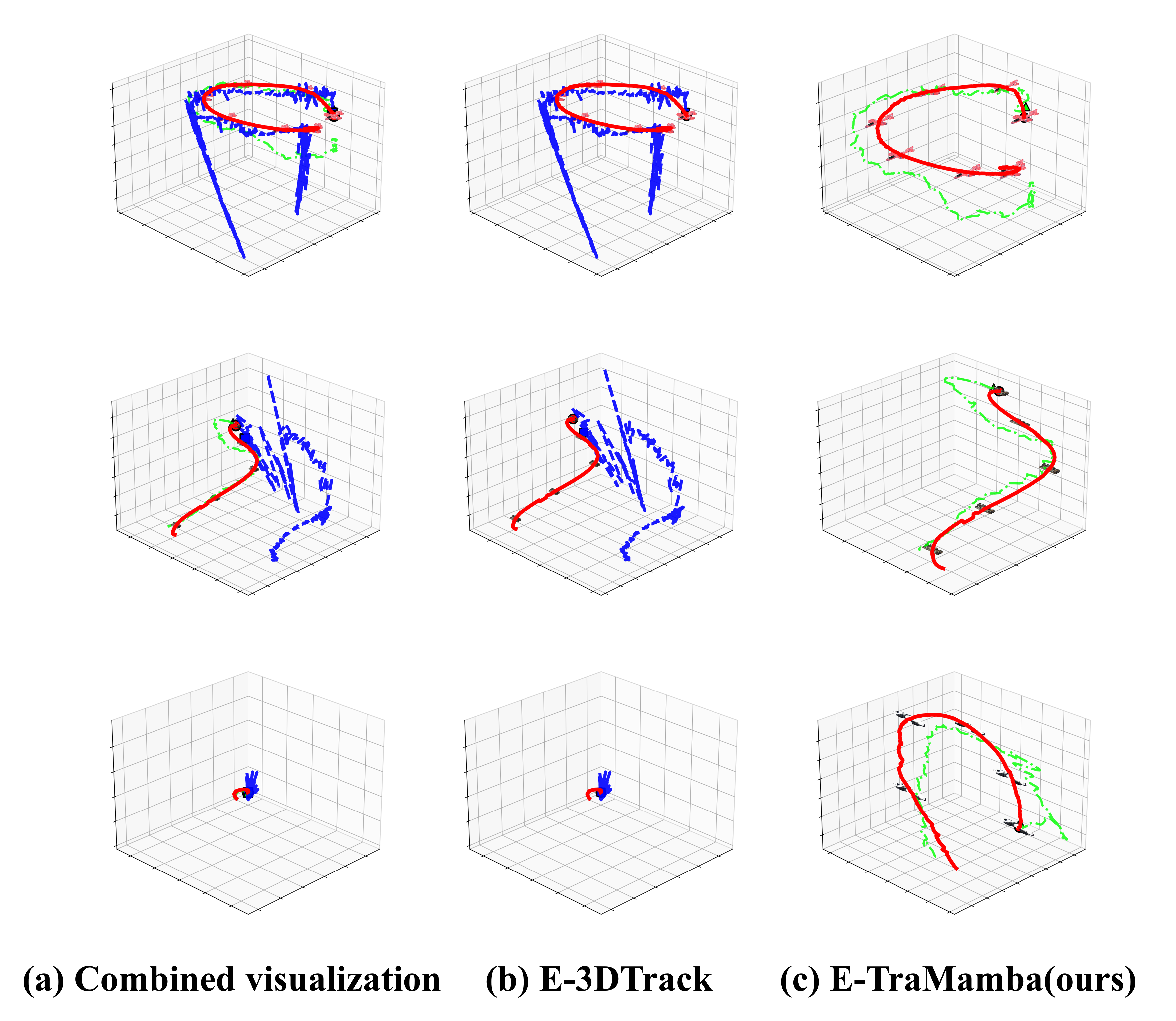}
    
    \caption{
    Comparison of 3D feature trajectories on the \textit{airplane}, \textit{toy1}, and \textit{toy5} sequences. 
    \textcolor{red}{Red:} ground-truth trajectories; 
    \textcolor{blue}{Blue:} E-3DTrack~\cite{3DEvpointtrac} baseline; 
    \textcolor{green!50!black}{Green:} E-TraMamba(our method). 
    Our model achieves smoother and more stable tracking across diverse 3D motion patterns with large depth and rotation changes.
    }
    
   \label{e3dtrack-vis}
\end{figure}

\subsection{Quantitative Comparison}

\paragraph{E-3DTrack results.}

Table~\ref{tab:combined_results_correct} presents the quantitative comparison of our method against existing event-based feature tracking and stereo depth estimation baselines on the challenging E-3DTrack dataset~\cite{3DEvpointtrac}. 
\textbf{E-TraMamba (ours)} consistently achieves the best performance across all metrics. 
Compared to the second-best approach, \textbf{E-3DTrack}~\cite{3DEvpointtrac}, 
our model reduces the RMSE by \textbf{22.6\%} (from 0.1181\,m to 0.0914\,m)
and improves feature age (FA) by \textbf{+35.4}, \textbf{+36.7}, and \textbf{+32.3} percentage points
at the 0.1\,m, 0.15\,m, and 0.2\,m thresholds, respectively.

These improvements stem from the synergy of two key modules: the spline-based motion estimator, which provides smooth local priors, and the Mamba-based state-space module, which refines long-range temporal dependencies with linear complexity $O(NT)$. This combination yields stronger temporal coherence and spatial stability under high-speed motion and depth variation, without incurring the quadratic cost of attention-based designs.

\paragraph{EvD-PointOdyssey results.}

Table~\ref{tab:combined_results_correct} further presents a quantitative comparison between our model and two representative event-based tracking frameworks on the EvD-PointOdyssey dataset. 
All methods are evaluated in the 2D image domain for fair comparison. 
\textbf{E-TraMamba (ours)} achieves the best performance on nearly all metrics: it performs on par with ETAP~\cite{hamann2025Evpointtrac} at the strictest FA@2px threshold (0.2177 vs.\ 0.2179) and outperforms both ETAP and DeepEvT~\cite{messikommer2023Evpointtrac} on all other metrics.
In particular, our model improves feature age (FA) at the 3\,px and 5\,px thresholds by \textbf{+10.4} and \textbf{+33.1} percentage points over ETAP,
and reduces the 2D RMSE from \textbf{23.26\,px} to only \textbf{4.16\,px}—a nearly \textbf{5.6$\times$} improvement. 
These results highlight that jointly modeling spatio-temporal motion and depth-conditioned event cues enables more stable and geometrically consistent long-term tracking under complex, high-speed event dynamics.

Despite the additional preprocessing and channel fusion required for event--depth inputs, 
our \textbf{E-TraMamba} architecture achieves \textbf{inference latency} nearly identical to lightweight 2D Transformer-based trackers. 
It operates at \textbf{2.11\,ms/frame} (\textbf{$\approx$475\,FPS}), matching DeepEvT (\textbf{2.10\,ms/frame}) and surpassing ETAP (\textbf{2.86\,ms/frame}).
This efficiency stems from the linear-time state-space modeling of the Mamba backbone, which replaces the quadratic attention mechanism in Transformers with recurrent state updates. 
As a result, E-TraMamba achieves real-time performance while incorporating richer 3D cues and spatio-temporal consistency—a combination that was previously challenging for Transformer-based designs.

\begin{figure}[t]
  \centering
   \includegraphics[width=\linewidth]{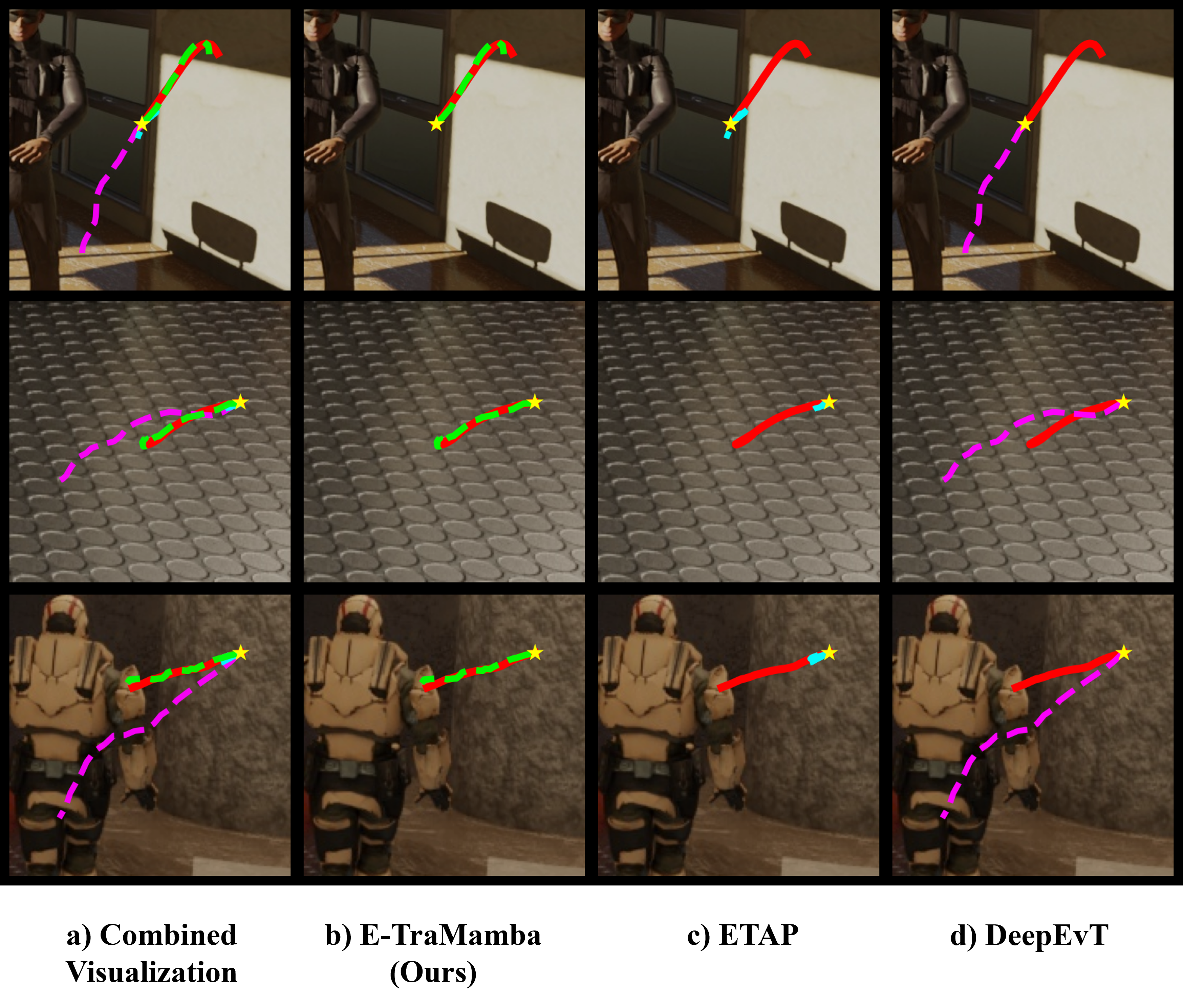}
    \caption{
    \textbf{Qualitative comparison of event-based 3D feature trajectories on challenging sequences.}
    \textcolor{red}{Red:} ground-truth trajectories; 
    \textcolor{green!60!black}{Green:} \textbf{E--TraMamba (ours)};
    \textcolor{cyan}{Cyan:} ETAP~\cite{hamann2025Evpointtrac};
    \textcolor{magenta}{Magenta:} DeepEvT~\cite{messikommer2023Evpointtrac};
    \textcolor{yellow!80!black}{Yellow~$\boldsymbol{\star}$:} initial feature locations. 
    \textbf{a)} Combined visualization of all methods. 
    \textbf{b)}–\textbf{d)} show individual results for each method, respectively. 
    Our approach produces smoother, longer, and more stable trajectories even under motion blur and complex texture regions.
    }
    \label{evdpo-vis}
\end{figure}

\subsection{Qualitative Comparison}

\paragraph{E-3DTrack results.}

Figure \ref{e3dtrack-vis} shows the qualitative 3D feature-tracking results of our proposed method compared with the prior event-based 3D tracking framework E-3DTrack \cite{3DEvpointtrac}.
The visualizations correspond to three representative sequences—airplane, toy1, and toy5—from our benchmark set, covering a wide range of motion patterns including large-scale translation, rotation, and depth variation.
Ground-truth trajectories are shown in red, those predicted by the comparison baseline in blue, and those from our method in green. From the figure, we observe that our method produces smoother and more stable 3D tracks, maintaining accurate temporal coherence even under fast motion and severe viewpoint changes.
In contrast, E-3DTrack suffers from noticeable jitter and drift, especially when the objects undergo complex depth changes or partial occlusion.
These results highlight the effectiveness of our motion-aware spatiotemporal modeling and the robustness of our long-range state representation in reconstructing continuous 3D trajectories.

\paragraph{EvD-PointOdyssey results.}

Figure~\ref{evdpo-vis} presents the qualitative 2D feature-tracking comparison on the \textbf{EvD-PointOdyssey} dataset. 
Each row shows a representative sequence exhibiting challenging motion patterns, including fast rotation, large depth variation, and complex textured regions. 
Ground-truth trajectories are shown in \textcolor{red}{red}, 
while our method \textbf{E--TraMamba}~(\textcolor{green!60!black}{green}) produces smoother and more stable long-term tracks than 
ETAP~\cite{hamann2025Evpointtrac}~(\textcolor{cyan}{cyan}) and DeepEvT~\cite{messikommer2023Evpointtrac}~(\textcolor{magenta}{magenta}),
particularly under severe motion blur and highly textured surfaces. 
\textbf{Star markers}~(\textcolor{yellow}{$\boldsymbol{\star}$}) denote the \textbf{initial feature locations}. 
These results highlight the effectiveness of our depth-conditioned event representation and the robustness of the Mamba-based long-range modeling in maintaining precise temporal correspondence across complex dynamic scenes.

\subsection{Ablation Studies}
To further analyze the effectiveness of individual components, we conduct comprehensive ablation studies summarized in Table~\ref{tab:ablation_structure} and Table~\ref{tab:ablation_modules}. 
We investigate three key factors: \textbf{(1) sequence modeling strategy (Transformer vs.\ Mamba)}, \textbf{(2) temporal window size}, and \textbf{(3) encoding configurations}.

\textbf{Impact of sequence modeling.}
As shown in Table~\ref{tab:ablation_structure}, when the window size is fixed at 8, the Transformer slightly outperforms Mamba (TFR$_{0.1}$: 0.5772 vs.\ 0.5661; RMSE: 0.0973 vs.\ 0.1006),
highlighting the strength of attention-based modeling.
However, Transformers suffer from quadratic complexity with respect to the window size ($\mathcal{O}(w^2)$), 
while Mamba maintains linear complexity ($\mathcal{O}(w)$). 
As the window increases to 12 and 16, Mamba surpasses Transformer both in accuracy and efficiency, 
making it more suitable for long-range and real-time event tracking tasks.

\begin{table}[t]
\centering
\small
\renewcommand{\arraystretch}{1.1}
\setlength{\tabcolsep}{4pt}
\begin{tabular}{c|c|c|c|c|c}
\toprule
Backbone & WinSize & Grps & TFR$_{0.1}$ $\uparrow$ & RMSE $\downarrow$ & FLOPs \\
\midrule
Transformer & 8 & 2 & 0.5772 & 0.0973 & 650G \\
Mamba       & 8 & 2 & 0.5661 & 0.1006 & 655G \\
Mamba       & 12 & 2 & \textbf{0.5950} & \textbf{0.0921} & 568G \\
Mamba       & 16 & 2 & 0.5896 & 0.0952 & \textbf{496G} \\
Mamba       & 12 & 3 & 0.5920 & 0.0935 & 571G \\
\bottomrule
\end{tabular}
\caption{Impact of backbone architecture, window size, and group count. FLOPs are computed for a 24-frame sequence with 20 tracked points, considering the sliding window strategy (stride = window\_size / 2).}
\label{tab:ablation_structure}
\end{table}

\begin{table}[t]
\centering
\small
\renewcommand{\arraystretch}{1.1}
\setlength{\tabcolsep}{6pt}
\begin{tabular}{c|c|c|c}
\toprule
PosEnc & TempEnc & TFR$_{0.1}$ $\uparrow$ & RMSE $\downarrow$ \\
\midrule
\checkmark & \checkmark & 0.5950 & 0.0921 \\
\checkmark & \xmark     & \textbf{0.6192} & \textbf{0.0911} \\
\xmark     & \xmark     & 0.5952 & 0.0960 \\
\bottomrule
\end{tabular}
\caption{Encoding ablation (win=12).}
\label{tab:ablation_modules}
\end{table}

\textbf{Effect of temporal window size.}
Window size plays a critical role in balancing context and stability. 
Although a longer window provides richer temporal information, our sliding strategy uses a stride of half the window size. 
Thus, overly large windows (e.g., 16) span excessive motion intervals, increasing temporal ambiguity and reducing point-wise precision. 
A window size of 12 achieves the best trade-off between temporal context, computational cost, and prediction stability.

\textbf{Encoding configurations.}
As shown in Table~\ref{tab:ablation_modules}, removing the temporal encoding slightly improves performance (TFR$_{0.1}$: 0.5950 $\rightarrow$ 0.6192), suggesting that the sequential scanning order of the Mamba modules already conveys sufficient temporal structure,
while additionally removing the positional encoding leads to a pronounced drop (TFR$_{0.1}$: 0.6192 $\rightarrow$ 0.5952; RMSE: 0.0911 $\rightarrow$ 0.0960).
This indicates that spatial cues are more critical than explicit temporal ordering in our event-based tracking setup.

\textbf{Number of Mamba groups.}
We further analyze the effect of stacking multiple Mamba refinement groups—each consisting of four directional Mamba blocks modeling forward/backward dependencies across spatial and temporal axes. 
Increasing the number of groups from 2 to 3 brings no further improvement (TFR$_{0.1}$: 0.5950 $\rightarrow$ 0.5920; RMSE: 0.0921 $\rightarrow$ 0.0935) while adding computational cost.
Therefore, we adopt two Mamba groups in the final model as a balanced configuration between accuracy and efficiency.

\section{Discussion}

\textbf{Efficiency, scalability, and performance of Mamba.}
Mamba achieves \textbf{linear-time temporal modeling} by reformulating feature propagation as a recurrent state-space process, 
avoiding the quadratic complexity of attention-based architectures. 
This efficiency is crucial for high-frequency event streams, 
where the number of effective \emph{virtual timesteps} reflects asynchronous sampling rather than real-world time. 
Under equivalent latency or FLOPs, Mamba processes longer temporal horizons without additional delay, 
enabling broader receptive fields and stronger temporal coherence. 
Consequently, \textbf{E-TraMamba} delivers higher accuracy and stability than Transformer-based trackers, 
particularly under motion blur and complex textures, while maintaining real-time performance. 
Experiments on the semi-synthetic \textbf{EvD-PointOdyssey} dataset further demonstrate strong cross-domain robustness 
to noise, domain shifts, and varying motion dynamics.

\textbf{Limitations and future directions.}
Despite these advantages, several challenges remain in transitioning from dense track-any-points setups to fine-grained, feature-level 3D tracking. 
In particular, \textbf{trajectory filtering, temporal consistency, and supervision reliability} remain difficult under real-world asynchronous conditions. 
Future work should explore decomposing large-scale tracking into interpretable sub-tasks that explicitly evaluate long-term modeling, 
temporal alignment, and efficiency under strict resource constraints. 
Incorporating physical priors or event-generation dynamics into Mamba-based architectures also presents a promising direction for further improvement.

\section{Conclusion}

We have presented \textbf{E-TraMamba}, a Mamba-based framework that establishes a new paradigm for efficient event-based 3D feature tracking. By reframing temporal modeling under a linear state-space formulation, E-TraMamba enables scalable, long-range reasoning with linear complexity and achieves stable performance where traditional attention or recurrent structures struggle.

Our work demonstrates that E-TraMamba achieves superior tracking accuracy and temporal consistency under high-speed motion and occlusion, at a lower computational cost than Transformer-based methods. These findings establish structured state-space modeling as a powerful and practical foundation for next-generation event-driven vision systems.

Our contribution extends beyond the model itself. We also present EvD-PointOdyssey, a high-fidelity semi-synthetic benchmark designed for unified 2D and 3D event tracking. Together, the model and dataset advance the frontier of event-based perception, emphasizing efficiency, scalability, and physical realism. Moving forward, we envision this combination as a foundation for real-time applications in autonomous navigation, agile robotics, and neuromorphic sensing—where low latency and long-term spatiotemporal understanding are essential.
{
    \small
    \bibliographystyle{ieeenat_fullname}
    \bibliography{main}
}
\setcounter{page}{1}
\maketitlesupplementary
\label{sec:appendix}

\section{Architecture Details}
\label{sec:appendix-arch}

This section provides complete implementation details for the \textbf{E-TraMamba} framework,
including precise network architectures, channel dimensions, and hyperparameters.
Our model supports both \textbf{stereo} and \textbf{monocular+depth} configurations,
which we describe separately below.

\subsection{Input Representations}
\label{sec:appendix-input}

\paragraph{Stereo Configuration.}
The input consists of synchronized left and right event streams,
each represented as a 10-channel tensor $E^{\text{L}}_t, E^{\text{R}}_t \in \mathbb{R}^{10\times H\times W}$,
where the 10 channels encode event polarity (positive/negative) across 5 temporal bins.
A reference grayscale patch $P^{\text{ref}}_i \in \mathbb{R}^{1\times 31\times 31}$
is extracted from the first frame for each tracked point.

\paragraph{Monocular+Depth Configuration.}
The input consists of a single event stream $E_t \in \mathbb{R}^{10\times H\times W}$
paired with a dense depth map $D_t \in \mathbb{R}^{1\times H\times W}$.
Depth values are normalized to $[0,1]$ via:
\[
D_t^{\text{norm}} = \frac{\text{clip}(D_t, d_{\min}, d_{\max}) - d_{\min}}{d_{\max} - d_{\min}},
\]
where $d_{\min}=0.9$\,m and $d_{\max}=50$\,m for PointOdyssey.
The normalized depth is concatenated with events to form an 11-channel input:
\[
I_t = [E_t,\, D_t^{\text{norm}}] \in \mathbb{R}^{11\times H\times W}.
\]

\subsection{Feature Extraction CNN}
\label{sec:appendix-cnn}

For each tracked point, we extract a $31{\times}31$ patch centered at its predicted 2D location.
The patch is processed by a CNN encoder:

\noindent\textbf{Stereo Mode:}
\begin{align*}
&\text{Conv2d}(10, 64, k=3, p=1) \to \text{BN}(64) \to \text{ReLU} \\
&\to \text{Conv2d}(64, 128, k=3, p=1) \to \text{BN}(128) \to \text{ReLU} \\
&\to \text{Conv2d}(128, 256, k=3, p=1) \to \text{BN}(256) \to \text{ReLU} \\
&\to \text{AdaptiveAvgPool2d}(1) \to F_{\text{local}} \in \mathbb{R}^{256}.
\end{align*}

\noindent\textbf{Mono+Depth Mode:}
\begin{align*}
&\text{Conv2d}(11, 64, k=3, p=1) \to \text{BN}(64) \to \text{ReLU} \\
&\to \text{Conv2d}(64, 128, k=3, p=1) \to \text{BN}(128) \to \text{ReLU} \\
&\to \text{Conv2d}(128, 256, k=3, p=1) \to \text{BN}(256) \to \text{ReLU} \\
&\to \text{AdaptiveAvgPool2d}(1) \to F_{\text{local}} \in \mathbb{R}^{256}.
\end{align*}

The reference patch encoder shares the same architecture but processes only the grayscale template $P^{\text{ref}}_i \in \mathbb{R}^{1\times 31\times 31}$:
\begin{align*}
&\text{Conv2d}(1, 64, k=3, p=1) \to \text{BN}(64) \to \text{ReLU} \\
&\to \text{Conv2d}(64, 128, k=3, p=1) \to \text{BN}(128) \to \text{ReLU} \\
&\to \text{Conv2d}(128, 256, k=3, p=1) \to \text{BN}(256) \to \text{ReLU} \\
&\to \text{AdaptiveAvgPool2d}(1) \to F_{\text{ref}} \in \mathbb{R}^{256}.
\end{align*}

\subsection{Correlation Features}
\label{sec:appendix-corr}

To capture temporal motion consistency, we compute correlation descriptors between consecutive frames.

\paragraph{Stereo Mode.}
For each tracked point at time $t$:
\[
\begin{aligned}
\Delta x_t,\;\Delta y_t &= (x_t - x_{t-1},\, y_t - y_{t-1}),\\
\rho_t &= \sum_{i,j,c} P_t^{\text{L}}(i,j,c)\,P_{t-1}^{\text{L}}(i,j,c),
\end{aligned}
\]
where $P_t^{\text{L}}$ and $P_{t-1}^{\text{L}}$ are $31{\times}31{\times}10$ event patches from the left view.
The concatenated vector $[\Delta x_t,\Delta y_t,\rho_t] \in \mathbb{R}^{3}$
is processed by an MLP:
\begin{align*}
&\text{Linear}(3, 128) \to \text{ReLU} \\
&\to \text{Linear}(128, 256) \to \text{ReLU} \\
&\to F_{\text{corr},t} \in \mathbb{R}^{256}.
\end{align*}

\paragraph{Mono+Depth Mode.}
We additionally include depth change:
\[
\Delta d_t = d_t - d_{t-1},
\]
yielding a 4-dimensional input $[\Delta x_t,\Delta y_t,\Delta d_t,\rho_t]$:
\begin{align*}
&\text{Linear}(4, 128) \to \text{ReLU} \\
&\to \text{Linear}(128, 256) \to \text{ReLU} \\
&\to F_{\text{corr},t} \in \mathbb{R}^{256}.
\end{align*}

\subsection{Position Encoding}
\label{sec:appendix-pos}

Normalized 2D coordinates $\left(\tfrac{x_t}{W},\tfrac{y_t}{H}\right)$
are encoded via:
\begin{align*}
&\text{Linear}(2, 128) \to \text{ReLU} \\
&\to \text{Linear}(128, 256) \to F_{\text{pos},t} \in \mathbb{R}^{256}.
\end{align*}

\subsection{Feature Fusion}
\label{sec:appendix-fusion}

\paragraph{Stereo Mode.}
Left and right view features are first extracted independently,
then fused via an MLP:
\begin{align*}
G^{i,\text{L}}_t &= F^{\text{L}}_{\text{local},t} + F_{\text{corr},t} + F_{\text{ref},t} + F_{\text{pos},t}, \\
G^{i,\text{R}}_t &= F^{\text{R}}_{\text{local},t} + F_{\text{corr},t} + F_{\text{ref},t} + F_{\text{pos},t}, \\
G^i_t &= \text{MLP}_{\text{stereo}}([G^{i,\text{L}}_t, G^{i,\text{R}}_t]),
\end{align*}
where $\text{MLP}_{\text{stereo}}$:
\begin{align*}
&\text{Linear}(512, 256) \to \text{ReLU} \\
&\to \text{Linear}(256, 256) \to G^i_t \in \mathbb{R}^{256}.
\end{align*}

\paragraph{Mono+Depth Mode.}
Features are concatenated and fused:
\begin{align*}
F_t^{\text{cat}} &= [F_{\text{local},t}, F_{\text{corr},t}, F_{\text{ref},t}] \in \mathbb{R}^{768}, \\
F_t^{\text{fuse}} &= \text{MLP}_{\text{fusion}}(F_t^{\text{cat}}) + F_{\text{pos},t},
\end{align*}
where $\text{MLP}_{\text{fusion}}$:
\begin{align*}
&\text{Linear}(768, 512) \to \text{ReLU} \\
&\to \text{Linear}(512, 256) \to F_t^{\text{fuse}} \in \mathbb{R}^{256}.
\end{align*}

\subsection{Mamba-based Refinement}
\label{sec:appendix-mamba}

The fused tokens $\mathbf{T}\!\in\!\mathbb{R}^{B \times T \times N \times 256}$
are refined through 2 groups of four-directional Mamba modules.

\paragraph{Mamba Block Configuration.}
Each Mamba module uses:
\begin{itemize}
\item \textbf{Hidden dimension}: $d = 256$
\item \textbf{State dimension}: $d_{\text{state}} = 32$
\item \textbf{Convolution kernel size}: $d_{\text{conv}} = 4$
\item \textbf{Expansion factor}: $\text{expand} = 2$
\end{itemize}

This yields an internal dimension of $d_{\text{inner}} = d \times \text{expand} = 512$.

\paragraph{Four-Directional Scanning.}
For each refinement group, tokens are scanned along four directions:
\begin{enumerate}
\item \textbf{Temporal forward} ($\tau^+$): reshape to $[B \cdot N, T, 256]$, process, reshape back.
\item \textbf{Spatial forward} ($\sigma^+$): reshape to $[B \cdot T, N, 256]$, process, reshape back.
\item \textbf{Temporal backward} ($\tau^-$): flip temporal axis, process, flip back.
\item \textbf{Spatial backward} ($\sigma^-$): flip spatial axis, process, flip back.
\end{enumerate}
Outputs are averaged: $\mathbf{T}' = \tfrac{1}{4}\sum_{s\in\mathcal{S}} \mathbf{H}_s$,
followed by residual connection: $\mathbf{T} \leftarrow \mathbf{T} + \mathbf{T}'$.

\subsection{Prediction Heads}
\label{sec:appendix-heads}

\paragraph{2D Position Heads.}
X and Y coordinates are predicted via softmax-weighted binning:
\begin{align*}
&\text{Linear}(256, 128) \to \text{ReLU} \\
&\to \text{Linear}(128, 31) \to \text{Softmax} \to \text{weighted sum} \to \Delta x,
\end{align*}
where the 31 bins span $\Delta x \in [-12.5, 12.5]$ pixels.
Y prediction follows the same structure.

\paragraph{Disparity Head (Stereo).}
\begin{align*}
&\text{Linear}(256, 128) \to \text{ReLU} \\
&\to \text{Linear}(128, 193) \to \text{Softmax} \to d \in [0, 192].
\end{align*}

\paragraph{Depth Head (Mono+Depth).}
\begin{align*}
&\text{Linear}(256, 128) \to \text{ReLU} \\
&\to \text{Linear}(128, 1) \to \Delta d,
\end{align*}
predicting depth increment $\Delta d$, followed by clipping to $[d_{\min}, d_{\max}]$.

\paragraph{Confidence Head.}
\begin{align*}
&\text{Linear}(256, 128) \to \text{ReLU} \\
&\to \text{Linear}(128, 1) \to \Delta c,
\end{align*}
predicting confidence increment $\Delta c$ in logit space.

\section{Dataset Construction Details}
\label{sec:appendix-dataset}

\subsection{Overview}
As an implementation supplement to Sec.~\ref{semi-syn},
we describe how long PointOdyssey sequences are segmented into
training windows for \textbf{EvD-PointOdyssey}.

Each scene is rendered into high-frame-rate RGB videos (240 FPS), converted to
events using a \emph{single, fixed} contrast threshold $\theta=0.02$
(calibrated once for the dataset), and paired with dense depth maps.
We do \emph{not} use multi-threshold synthesis, ensuring consistent event generation
across the entire dataset.
The resulting event--depth--trajectory triplets are then organized
into windowed training sequences with controlled motion magnitude.

\subsection{Sequence Formation}
From the filtered trajectories in Sec.~\ref{semi-syn},
we extract sliding temporal windows of length
$\ell \in \{12,16,24,30\}$ frames.
A window is valid if it satisfies:
\begin{enumerate}
\item \textbf{Full visibility}: all selected points satisfy $v_{t,n}=1$ for all $t\in [s,s+\ell)$,
\item \textbf{Non-overlapping}: frame ranges do not overlap with previously selected windows,
\item \textbf{Motion constraints}: maximum inter-frame displacement $\Delta_{n}^{\max} \leq 12$\,px
and mean displacement $\Delta_{n}^{\text{avg}} \geq 2$\,px,
\item \textbf{Depth validity}: all depth values are finite and within $[0.9, 50]$ meters,
\item \textbf{Point count}: $1 \leq N \leq 30$ valid points per sequence.
\end{enumerate}

Each valid sequence contains:
\begin{itemize}
\item Per-frame 10-channel event tensors $E_t \in \mathbb{R}^{10\times H\times W}$
(5 positive \& 5 negative temporal bins),
\item Dense depth maps $D_t \in \mathbb{R}^{1\times H\times W}$,
\item 2D pixel trajectories $\mathbf{X}_{t,n} \in \mathbb{R}^{2}$,
\item 3D world trajectories $\mathbf{P}_{t,n} \in \mathbb{R}^{3}$,
\item Visibility masks $v_{t,n} \in \{0,1\}$,
\item Camera intrinsics and poses.
\end{itemize}

This segmentation strategy supports both short-term
($12$–$16$ frames) and longer-term ($24$–$30$ frames) motion patterns,
enabling the model to learn across different temporal scales.

\subsection{Filtering Algorithm}
Algorithm~\ref{alg:sequence} summarizes the selection procedure used
to generate valid training sequences from the PointOdyssey assets.

\begin{algorithm}[h]
\caption{Smart sequence selection for \textbf{EvD-PointOdyssey}.}
\label{alg:sequence}
\small
\begin{flushleft}
1:\ \textbf{Input:} 2D trajectories $\mathbf{X}_{t,n}$, visibility masks $v_{t,n}$, depth maps $d_{t,n}$\\
2:\ \textbf{for} each start frame $s$ \textbf{do}\\
3:\ \hspace*{1.2em}\textbf{for} each length $\ell \in \{30,24,16,12\}$ \textbf{do} \hfill $\triangleright$ prioritize longer sequences\\
4:\ \hspace*{2.4em}define window $[s,s+\ell)$; skip if overlapping with existing sequences\\
5:\ \hspace*{2.4em}collect points with $v_{t,n}=1$ for all $t\in [s,s+\ell)$\\
6:\ \hspace*{2.4em}compute max displacement $\Delta_{n}^{\max}=\max_{t}\|\mathbf{X}_{t+1,n}-\mathbf{X}_{t,n}\|$\\
7:\ \hspace*{2.4em}compute mean displacement $\Delta_{n}^{\text{avg}}=\frac{1}{\ell-1}\sum_{t}\|\mathbf{X}_{t+1,n}-\mathbf{X}_{t,n}\|$\\
8:\ \hspace*{2.4em}keep points with $2 \le \Delta_{n}^{\text{avg}}\le 12$\,px and finite depth\\
9:\ \hspace*{2.4em}\textbf{if} valid points $1\le N\le30$ \textbf{then}\\
10:\hspace*{3.6em}record sequence meta-data; mark frames $[s,s+\ell)$ as used\\
11:\hspace*{3.6em}\textbf{break} \hfill $\triangleright$ proceed to next start frame\\
12:\hspace*{2.4em}\textbf{end if}\\
13:\hspace*{1.2em}\textbf{end for}\\
14:\ \textbf{end for}\\
15:\ \textbf{Output:} list of valid sequences with point IDs, events, depth, and poses
\end{flushleft}
\end{algorithm}

\subsection{Dataset Statistics}
The final \textbf{EvD-PointOdyssey} dataset contains:
\begin{itemize}
\item \textbf{Training}: 2,847 sequences from 100 unique scenes
\item \textbf{Testing}: 312 sequences from 11 held-out scenes
\item \textbf{Length distribution}: 30-frame (42\%), 24-frame (28\%), 16-frame (18\%), 12-frame (12\%)
\item \textbf{Average points per sequence}: 18.3
\item \textbf{Average displacement}: 4.2 pixels/frame
\item \textbf{Average depth range}: 1.2–28.5 meters
\end{itemize}

\subsection{Reproducibility Assets}
Complete per-scene sequence lists, start/end indices, and detailed
statistics (sequence lengths, point counts, displacement distributions)
will be released together with the dataset.
These assets enumerate the exact training windows used in our experiments,
enabling full reproducibility of our dataset construction pipeline.


\end{document}